\def\year{2022}\relax
\documentclass[letterpaper]{article} 
\usepackage{aaai22}  
\usepackage{times}  
\usepackage{helvet} 
\usepackage{courier}  
\usepackage[hyphens]{url}  
\usepackage{graphicx} 
\def\UrlFont{\rm}  
\usepackage{natbib}  
\usepackage{caption} 
\usepackage{times}
\usepackage{epsfig}
\usepackage{graphicx}
\usepackage{amsmath}
\usepackage{amssymb}
\usepackage{caption}
\usepackage{booktabs}       
\usepackage{amsfonts}       
\usepackage{nicefrac}       
\usepackage{microtype}      
\usepackage{mathtools}
\usepackage{arydshln}
\usepackage{xspace}
\usepackage{bm}

\usepackage{xcolor,colortbl}
\definecolor{Gray}{gray}{0.9}
\usepackage{soul}
\DeclareRobustCommand{\hlgray}[1]{{\sethlcolor{Gray}\hl{#1}}}

\usepackage{algorithmicx}
\usepackage{algorithm}
\usepackage[noend]{algpseudocode}
\usepackage{pifont}
\usepackage{etoolbox}\AtBeginEnvironment{algorithmic}{\footnotesize}

\usepackage{arydshln}
\usepackage{amsmath}
\usepackage{multirow}

\usepackage{tikz}
\usetikzlibrary{matrix,chains,decorations.pathreplacing,fit,calc,quotes}
\usetikzlibrary{patterns}
\usetikzlibrary{positioning,automata}

\usepackage{listings}
\usepackage{soul}

\definecolor{scheme-primitive}{rgb}{0, 0, 1}
\definecolor{scheme-derived}{rgb}{0, 0, 1}
\definecolor{scheme-comment}{rgb}{0, 0.5, 0}
\definecolor{scheme-bracket}{rgb}{0.6, 0.6, 0.6}

\lstdefinelanguage{lispMore}{stepnumber=1,
firstnumber=1,
numbersep=5pt,language=Lisp,
morekeywords={if,=},
stringstyle=\ttfamily,
basicstyle=\footnotesize\ttfamily, 
showstringspaces=false,
breaklines=false,
stringstyle=\color{scheme-comment},
keywordstyle=\color{scheme-derived},
moredelim=**[is][\bfseries\color{red}]{@}{@},
moredelim=**[is][\bfseries\color{purple!40!black}]{!}{!},
frame=single
}

\usepackage[T1]{fontenc}

\usepackage[utf8]{inputenc}

\usepackage{microtype}

\title{Pushing the Limits of Rule Reasoning in Transformers\\ through Natural Language Satisfiability}

\author{Kyle Richardson\ \ \ \ \ \ Ashish Sabharwal 
}
\affiliations{
Allen Institute for AI, Seattle, WA, USA \\
\{kyler,ashishs\}@allenai.org
}

\begin{document}

\maketitle

\begin{abstract}
Investigating the reasoning abilities of transformer models, and discovering new challenging tasks for them, has been a topic of much interest. Recent studies have found these models to be surprisingly strong at performing deductive reasoning over formal logical theories expressed in natural language. A shortcoming of these studies, however, is that they do not take into account that logical theories, when sampled uniformly at random, do not necessarily lead to hard instances. We propose a new methodology for creating challenging algorithmic reasoning datasets that focus on \emph{natural language satisfiability} (NLSat) problems.\footnote{Our data and code are available at \url{https://github.com/allenai/language_fragments}.} The key idea is to draw insights from empirical sampling of hard propositional SAT problems and from complexity-theoretic studies of language. This methodology allows us to distinguish easy from hard instances, and to systematically increase the complexity of existing reasoning benchmarks such as RuleTaker. We find that current transformers, given sufficient training data, are surprisingly robust at solving the resulting NLSat problems of substantially increased difficulty. They also exhibit some degree of \emph{scale-invariance}---the ability to generalize to problems of larger size and scope. Our results, however, reveal important limitations too: a careful sampling of training data is crucial for building models that generalize to larger problems, and transformer models' limited scale-invariance suggests they are far from learning robust deductive reasoning algorithms.
\end{abstract}

\section{Introduction}

Motivated by the impressive performance of recent pre-trained transformers \cite{devlin2018bert,raffel2019exploring} on a wide range of natural language understanding (NLU) benchmarks \cite{wang2019glue,wang2019superglue,xu-etal-2020-clue}, there has much been recent interest in investigating the linguistic and reasoning abilities of state-of-the-art neural models \citep[][\emph{inter alia}]{linzen2016assessing,talmor2019olmpics,kassner2020pre,yanaka2020neural,hupkes2020compositionality,richardson2020probing}. One particular thread of work focuses on probing whether transformers can perform logical reasoning over formal theories expressed in natural language \cite{clark2020transformers}. Specifically, given a set of systematically constructed \emph{natural language theories} consisting of a set of explicitly stated rules and facts (e.g., the \textbf{NL Theory} in the bottom part of Figure~\ref{fig:top_fig} containing fictional rules about characters \emph{Bob} and \emph{Alan}), the goal is to see whether a model can learn to perform deductive reasoning over such theories by correctly answering queries that require making novel inferences (e.g., predicating that \emph{Alan is green} is true based on knowing that \textbf{Alan is rough} and applying the rule \hlgray{All rough people are green}).

\begin{figure}
    \centering
    
    \begin{tikzpicture}[scale=0.7, every node/.style={scale=0.7}]
\tikzstyle{source} = [draw,thick,inner sep=.2cm,fill=Gray];
\tikzstyle{source2} = [draw,thick,inner sep=.2cm];
\tikzstyle{process} = [draw,thick,circle]; 
\tikzstyle{to} = [-,shorten >=1pt,semithick,font=\sffamily\footnotesize];
\matrix (compiler) 
[matrix of nodes,
  ampersand replacement=\&,
  column sep=.4cm,
  nodes={align=left},
  every node/.style={text width=35mm,text badly centered,text height=3mm,font=\scriptsize}
  ]
{
  \node[](dummy1){};  \& \node[](dummy2){}; \\[.1cm]
  \node[source2](result){
	\textbf{Hard Combinatorial Problem $\mathcal{P}$} {\scriptsize (e.g., SAT, 3-Coloring)}
  };  \& \node[source](challenge) {\textbf{Natural Language Probing Task $\mathcal{T}$} {\scriptsize (e.g., deductive inference, syntactic processing,...)}}; \\[.1cm] 
    \node[](performance){
  }; \& \node[](dummy5){}; \\[-.2cm] 
}; 
\draw[-,thick,draw=black] (challenge) -- node[above left] {} (dummy5.center);
\draw[-,thick,draw=black] (dummy5.center) -- node[below] {\emph{Find fragments of $\mathcal{T}$ grounded in $\mathcal{P}$}} (performance.center);
\draw[->,thick,draw=black] (performance.center) -- node[above left] {} (result.south);
\draw[-,thick,dashed,draw=red] (result.north) -- node[above left] {} (dummy1.center);
\draw[-,thick,dashed,draw=red] (dummy1.center) -- node[above] {\emph{Sample hard instances of $\mathcal{P}$}} (dummy2.center);
\draw[->,thick,dashed,draw=red] (dummy2.center) -- node[above] {} (challenge.north);
\end{tikzpicture}
\vspace{1ex}
    
    {\footnotesize
        \begin{tabular}{| c| p{5.5cm} |}
            \hline 
             \textbf{NL Theory} & {$\Gamma =$\{ \textit{Bob is round}. \textit{Alan is blue, rough and young}. \hlgray{If someone is round then they are big}. \hlgray{All rough people are green}. \hlgray{Big people are not green}. \}}  \\ \hline 
             \textbf{Conjectures} & 
             \begin{tabular}{l}
                \textbf{1.} Alan is green (\textbf{entailment}, $\Gamma \models \textbf{1}$) \\ 
                \textbf{2.} Bob is green (\textbf{contradiction}) 
             \end{tabular}
             \\ \hline 
             \textbf{Satisfiability} &  \begin{tabular}{l}
             $\Gamma$ has an interpretation (\textbf{sat}) \\ $\Gamma \cup \{ \neg \textbf{1} \}$ (\textbf{unsat}), \emph{indirectly proves} $\Gamma \models \textbf{1}$ \\ $\Gamma \cup \{ \textbf{2} \}$ (\textbf{unsat}), \emph{indirectly proves} $\Gamma \models \neg\textbf{2}$
             \end{tabular}
             \\ \hline 
        \end{tabular}}

    \caption{TOP: An illustration of our general methodology for constructing hard natural language reasoning problems for a task $\mathcal{T}$, by grounding them into a hard combinatorial problem $\mathcal{P}$ and sampling hard instances of $\mathcal{P}$. BOTTOM: An example of a \emph{natural language (NL) theory} (i.e., set of arbitrary \textit{facts} and \hlgray{rules}) $\Gamma$ along with two example \emph{conjectures} (i.e., propositions to be proved) and the relationship between entailment and satisfiability.}
    \label{fig:top_fig}
\end{figure}

While much of this recent work on \emph{behavioral probing} has centered around small synthetic domains and datasets  (see also \citet{weston2015towards,lake2018generalization,sinha2019clutrr}), the appeal of such testing is that it can allow us to uncover the strengths and weaknesses of models in a cost-effective and controlled manner, and ultimately determine whether models are inherently capable of solving certain algorithmic problems. Given that most behavioral probing studies are performed in a \emph{black-box} fashion \cite{ribeiro2020beyond} and are thus limited to input-output-driven testing, however, the quality and informativeness of a probing study relies on having reliable data that faithfully captures the full target problem space. In particular, to demonstrate that a model can learn a certain algorithmic skill, it must be demonstrated that the model can solve the hardest instances of the target problem. Indeed, recent work \cite{shin2019synthetic,wu2021reascan,tamari2021dyna} has revealed various pitfalls associated with synthetic data due to ad-hoc sampling strategies, which can dramatically inflate model performance by under-sampling difficult cases in a way that can also harm model generalization.


In evaluating a particular diagnostic dataset for probing logical reasoning, the following question arises: \emph{are the problems contained in the target dataset hard in some objective computational sense?} For example, while knowing that \emph{Bob is green} is false in Figure~\ref{fig:top_fig} requires making multiple inferential steps (i.e., combining the fact \textit{Bob is round} with the two rules \hlgray{If someone is round then they are big} and  \hlgray{Big people are not green}), the structure of the rules involved is such that there are well-known highly efficient algorithms for computing this inference.\footnote{More technically, such a query can be answered via a linear-time process called \emph{unit propagation} (cf.~\citet{zhang1996cient}), which is often treated as a pre-processing step in many modern theorem provers and SAT solvers.} A natural question, then, is: can models perform inferences involving more complex reasoning with rules? Answering the hardness question, therefore, involves two additional questions: \textbf{(Q1)} is the \emph{formal language} used to express the target problem space capable of expressing \emph{hard} problems (e.g., ones that go beyond simple linear chaining)? \textbf{(Q2)} is the \emph{sampling method} used to generate target instances able to effectively capture the full problem space? 

In this paper, we fix the formal language to be expressive enough such that it can, by design, represent computationally hard problems (thereby addressing Q1). To address Q2, we propose a \textbf{general methodology}, illustrated in the top part of Figure~\ref{fig:top_fig}. Given a target probing task $\mathcal{T}$ such as deductive inference over statements expressed in natural language, the key idea is to identify subsets of $\mathcal{T}$ that map to a known hard combinatorial reasoning problem $\mathcal{P}$ such as Boolean satisfiability (SAT), and devise methods to sample hard instances of $\mathcal{P}$ in order to arrive at hard instances of $\mathcal{T}$.

Specifically, we broaden the scope of \citet{clark2020transformers} to look at \emph{natural language satisfiability} (NLSat) problems, or types of natural language deductive reasoning problems that formally assume an underlying SAT semantics. Using insights from empirical sampling of hard SAT problems \cite{selman1996generating}  we show how to systematically construct computationally difficult reasoning problems by focusing on such hard rule fragments and by sampling from the critical phase-change regions of SAT. We show that such an approach has twofold utility: 1) distinguishing \emph{easy} from \emph{hard} instances that are consequential for training robust models and for reliable evaluation and; 2) for diagnosing and increasing the complexity of existing reasoning benchmarks. 

Our results are partly positive: when provided with a sufficient amount of training instances (e.g., >100k examples), recent pretrained transformers can indeed solve non-trivial NLSat problems that far exceed the complexity of existing reasoning benchmarks (e.g., achieving >90\% accuracy on quantified rule theories containing up to $70$ ground variables). They also exhibit some degree of generalization and \emph{scale-invariance}, or the ability to generalize to problems of larger scope (e.g., generalizing from propositional theories with $12$ variables to ones with $30$, while maintaining performance far above random chance).

At the same time, our results also reveal important caveats:
1) the ability of a model to solve hard reasoning problems critically relies on how well its training data is sampled and; 2) the degree to which models are scale-invariant remains limited, suggesting that models trained in the standard paradigm are still far from learning the underlying algorithms needed for robust deductive reasoning.

\section{Related Work}

Our work follows the literature on behavioral testing of neural NLU models and builds on work by \citet{clark2020transformers} on probing deductive reasoning, which has spawned a number of subsequent studies \cite{saha2020prover,gontier2020measuring,betz2020critical,betz2021thinking,tafjord2020proofwriter,saparov2021generative,liang2021explainable}. While these studies demonstrate that models are able to solve some deductive reasoning problems, we observe that existing datasets narrowly focus on the simplest deductive reasoning problems when subjected to closer analysis. As we detail in Table~\ref{tab:data_comparison}, the standard RuleTaker dataset, which is based on a fragment of English that is \emph{capable} of expressing intractably hard algorithmic problems, is limited to the easiest types of deductive reasoning problems. As a result, existing models lack robustness when evaluated on harder parts of the problem distribution as we show in Table~\ref{tab:hard-RT} on a RuleTaker-style dataset sampled using our new sampling strategy.


To find hard reasoning fragments of natural language, we take inspiration from the literature of complexity-theoretic studies of various natural language fragments 
\cite{pratt2004fragments,pratt2006more,pratt2009logics,thorne2010data}. Particularly, \citet{pratt2004fragments} looks at the computational properties such as the complexity of satisfiability for various rule fragments of English, which is the motivation behind the grounded relative-clause fragment we describe in the next section. While this work focuses on a worst-case analysis of different linguistic phenomena, we use the results as a guide to find the \emph{hard} cases for probing the limits of  models. 

\begin{figure}[t]
    \centering
    \includegraphics[scale=0.35]{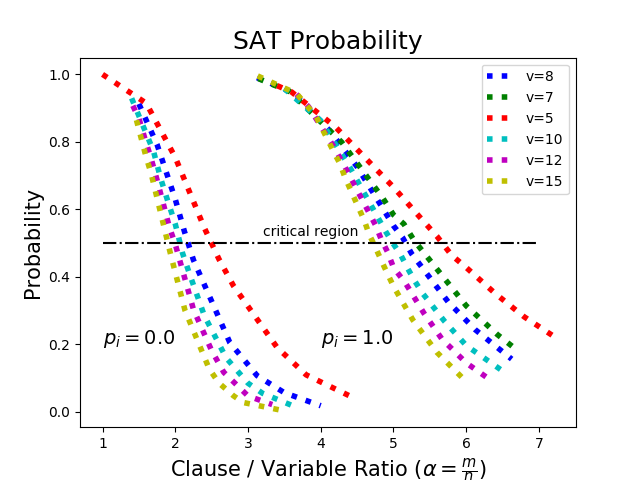}
    \caption{Illustration of phase-change and SAT probability for random $2\textsc{-SAT} (p_{\textrm{int}} = 0)$ and $3\textsc{-SAT} (p_{\textrm{int}}= 1)$ problems over a randomly sampled set of examples with varying $\alpha$ and number of variables (5-15).}
    \label{fig:phase_trans}
\end{figure}

To find hard natural language satisfiability instances, we rely on techniques from the literature on empirical sampling of combinatorial problems, where it has been observed \cite{cheeseman1991really} that hard instances of different problems lie at various critical thresholds that correlate with the \emph{constrainedness} of a given problem. We specifically use techniques for generating hard 3SAT problems  \cite{selman1996generating,cook1997finding} to sample hard problems from the critical regions of SAT phase transitions (see Figure~\ref{fig:phase_trans}). To our knowledge, we are first to investigate this work and using SAT-based representations of linguistic problems to empirically find hard natural language reasoning problems (see \citet{hahn2021sensitivity}).

Our study also follows other work on training neural models to solve hard algorithmic problems \cite{vinyals2015pointer,reed2015neural,cai2017making}, including SAT \cite{selsam2018learning} and propositional inference \cite{evans2018can,traylor2021and}; a key difference is our focus on algorithmic problems in natural language and on probing current pre-trained transformers \cite{devlin2018bert,liu2019roberta,raffel2019exploring}. Following studies such as \citet{reed2015neural}, we also look at the ability of models to be \emph{scale-invariant}, or scale to problems of larger size and scope. Within this space, we follow \citet{shin2019synthetic,wu2021reascan} in developing novel sampling strategies for avoiding the pitfalls of randomly sampled algorithmic datasets, which can give rise to the kinds of biases observed in human-annotated datasets \cite{gururangan2018annotation} and limit model generalization.

\begin{algorithm}[t]
\algnewcommand\algorithmicinput{\textbf{Input:}}
\algnewcommand\INPUT{\item[\algorithmicinput]}
\algnewcommand\algorithmicoutput{\textbf{Output:}}
\algrenewcommand\algorithmicindent{1em}
\algnewcommand\OUTPUT{\item[\algorithmicoutput]}
\caption{Dataset construction via random SAT}
\begin{algorithmic}[1]
\INPUT  Variables set $V = \{v_{1},...v_{n}\}$ of size $n$, natural language templates $\mathcal{R}$ and variables $\mathcal{P}$, 2SAT to 3SAT interpolation parameter $p_{\textrm{int}}$, negation parameter $p_{\textrm{neg}}$, clause variable ratio $\alpha$ range $(\alpha_{\min}, \alpha_{\max})$, \textsc{STOP}
condition 
\OUTPUT NLSat dataset

\State $\mathbf{D} \leftarrow \{ \}$\Comment{initialize dataset}
\Repeat
\State $P \leftarrow \{ \}$ \Comment{problem/clause set}
\State $m \sim $\textbf{ choose} $m$, s.t. $\alpha_{\min} \leq \frac{m}{n} \leq \alpha_{\max}$ 
\For{$i:=1 \textbf{ to } m$} \Comment{generate $m$ clauses}
\State $s \sim$ \textbf{choose} clause size $k \in$ \textcolor{red}{(3,2) with prob.\ $(p_{\textrm{int}}, 1-p_{\textrm{int}})$}
\State $\mathbf{V'} \sim $\textbf{ choose} $s$ unique variables from $V$
\State $\mathbf{C} \leftarrow$ \textbf{negate} each $v \in V'$ with $p_{\textrm{neg}}$ \Comment{new clause}
\State \textcolor{red}{$t \sim $\textbf{ choose} NL template from $\mathcal{R}$ of size $s$}
\State \textcolor{red}{$d \leftarrow $\textbf{ instantiate} $t$ over $\mathbf{C}$ using variables from $\mathcal{P}$}
\State $\mathbf{P} \leftarrow \mathbf{P} \cup \{ d \}$
\EndFor
\State $\mathbf{D} \leftarrow \mathbf{D} \cup P$
\Until {dataset \textsc{STOP} condition is met}
\end{algorithmic}
\label{alg:sampling}
\end{algorithm}

\section{Dataset Construction and Methodology}

Natural language satisfiability (NLSat) is a deductive reasoning task that involves determining whether a set of rules expressed in language (e.g., those shown in Figure~\ref{fig:lang_examples}) have a satisfying assignment or \emph{possible} interpretation.\footnote{As we show in the lower part of Figure~\ref{fig:top_fig}, logical entailment and satisfiability (or its complement) end up being intereducible notions for the logics under consideration. For basic results on the connection, see \citet{davis1994computability}[Theorem2.1,p252].} Following our general methodology shown in the top part of Figure~\ref{fig:top_fig}, in order to find \emph{hard} deductive reasoning problems of this kind, we sample hard instances from ordinary SAT problems in Boolean logic and translate them into natural language using a pre-defined set of English rule languages.

In this section, we first detail the semantics of SAT and how to identify hard SAT problems (\textbf{Identifying Hard Problems} and Algorithm~\ref{alg:sampling}), and then describe the two different fragments of English we use for our experiments (the \textbf{Grounded Rule Language} and \textbf{Grounded Relative Clause Fragment}; see details in Figure~\ref{fig:lang_description});  both borrow certain \emph{grounded} and \emph{quantified} rule constructs from the RuleTaker language and, in the latter fragment, build on some constructions studied in formal linguistics. Finally, we discuss ways of sampling SAT instances of different hardness levels and sizes.

\subsection{Identifying Hard Problems}
\label{sec:sampling}

The SAT problem is the classic NP-complete problem~\cite{cook1971complexity}. We focus on \textsc{3sat} problems where $k=3$, i.e., each formula is limited to clauses of size three. While \textsc{3sat} is computationally hard under a worst-case analysis, this does not mean that all, or even most, \textsc{3sat} instances are hard to solve. Indeed, work on empirical sampling of classes of random $k\textsc{-SAT}$ problems has revealed that whether a problem is difficult crucially relies on details about the target distribution from which the problems are sampled \cite{selman1996generating,mitchell1996some} as well as the particular parameters employed during sampling.

\begin{figure*}[t]

\centering 
{
\begin{tabular}{| p{5cm} p{10cm} |} 
\hline 
{ \textbf{Grounded Rule Language}} (GRL) & If (no) \hlgray{$\mathbf{X}$} and (no) \hlgray{$\mathbf{Y}$} then (not) \hlgray{$\mathbf{Z}$}.  \\ \hline
\textbf{Grounded Relative Clause Fragment} (RCL)  & Every \hlgray{$\mathbf{X}$} who is (not) a/an \hlgray{$\mathbf{Y}$} is (not) a/an \hlgray{$\mathbf{Z}$}. No \hlgray{$\mathbf{X}$} who is (not) a/an \hlgray{$\mathbf{Y}$} is a/an \hlgray{$\mathbf{Z}$}. Everyone who is (not) a/an \hlgray{$\mathbf{X}$} and (not) a/an \hlgray{$\mathbf{Y}$} is (not) a/an \hlgray{$\mathbf{Z}$}. \hlgray{$\mathbf{c}$} is (not) a/an \hlgray{$\mathbf{X}$} or (not) a/an \hlgray{$\mathbf{Y}$} or (not) a/an \hlgray{$\mathbf{Z}$}.  \\ \hline 
\end{tabular}}

\caption{A syntactic description of the two rule languages used for our experiments. }
\label{fig:lang_description}
\end{figure*}

 \begin{figure*}[t]
    \centering
    
    {
    \begin{tabular}{ | p{3cm} p{7.5cm} p{4.5cm} |}
        \hline 
        \textbf{Language} & \multicolumn{1}{c}{\textbf{Example Expression}} & \multicolumn{1}{c|}{\textbf{Satisfying Assignment}} \\ \hline 
        \textbf{Propositional Logic (3SAT)} & $(\neg \mathbf{v}_{1} \lor \mathbf{v}_{15} \lor \mathbf{v}_{13} ) \land  (  \neg \mathbf{v}_{13} \lor \neg \mathbf{v}_{12} \lor \neg \mathbf{v}_{1} ) \land  ( \mathbf{v}_{1} \lor \mathbf{v}_{15} \lor \neg \mathbf{v}_{13} ) \land ...$ & $v_{1}$=\textbf{false}, $v_{15}$=\textbf{false}, $v_{13}$=\textbf{false}, $v_{12}$=\textbf{true}...  \\ 
        \multicolumn{3}{|c|}{\textbf{Natural Language Fragments}} \\ \cdashline{1-3} 
        \textbf{Grounded} \newline \textbf{Rule} \newline \textbf{Language} (GRL) & If \hlgray{carrot} and not \hlgray{steak} then \hlgray{apples}, If \hlgray{apples} and \hlgray{grapes} then no \hlgray{carrots}. If no \hlgray{carrots} and no \hlgray{steak} then not \hlgray{apples}... & \textbf{needed:} \emph{carrots, apples, grapes},... \\ \hline 
        \textbf{Relative} \newline \textbf{Clause} \newline \textbf{Fragment} (RCL) & Every \hlgray{doctor} who is not a \hlgray{philosopher} is a \hlgray{baker}. No \hlgray{baker} who is a \hlgray{gardener} is a \hlgray{philosopher}. John is a \hlgray{doctor} or a \hlgray{philosopher} or not a \hlgray{baker}... & John can be a \textbf{doctor}, a \textbf{baker}, not a \textbf{philosopher} and not a \textbf{gardener}...  \\ \hline 
    \end{tabular}} 
    \caption{Example translations of a satisfiable 3SAT problem (truncated) in boolean logic and two fragments of English (variables in the natural language are \hlgray{highlighted}). The bottom shows example interpretations of each expression that demonstrate satisfiability.}
    \label{fig:lang_examples}
\end{figure*}
 
To obtain hard SAT instances, we rely on a variant of the well-studied random $k\textsc{-SAT}$ algorithm first introduced in \citet{selman1996generating}, which we illustrate in Algorithm~\ref{alg:sampling}. In standard $k\textsc{-sat}$, random formulae of size $m$ containing $n$ variables and clauses of fixed length $k$ are obtained by selecting $m$ clauses (starting \textbf{line 5}) uniformly from the space of $2^{k} \binom{n}{k}$ possible clauses (where each clause is constructed by sampling $k$ unique variables (\textbf{line 7}) and negating each with probability $p_{\textrm{neg}}$ (\textbf{line 8})). While our primary focus is on $3\textsc{-sat}$, for convenience we include the possibility of sampling mixed $2\textsc{-sat}$/$3\textsc{-sat}$ problems by introducing an \emph{interpolation} parameter $p_{\textrm{int}}$ (shown on \textbf{line 6}, \cite{monasson1999determining}). Using a suitable fragment of natural language $\mathcal{R}$ (see next section), our version additionally includes translating each random clause to expressions in natural language (\textbf{lines 9-10}).


A key parameter in random SAT is the clause to variable ratio $\alpha = \frac{m}{n}$ (computed on \textbf{line 4} and dictated, in part, by the range $\alpha_{\min},\alpha_{\max}$). This parameter gives rise to phase-change behavior that has implications for problem hardness \cite{hayes2003computing}. Such phase changes are illustrated in Figure~\ref{fig:phase_trans}, where $\alpha$ (x-axis) can be used to determine the probability of a random formula being satisfiable (y-axis). For our purposes, such a curve suggests a principled way to identify \emph{hard} instances, namely, by selecting formulae from the \emph{critical region} where problems have roughly $0.5$ probability of being satisfiable. The motivation behind sampling in this manner follows much of the work in empirical SAT, where is it found that such problems are constrained in a unique way that makes it difficult to simply \emph{guess} the correct answer by looking at the superficial patterns, which in our context makes it harder for model to exploit short-cuts (we later provide empirical evidence that narrowly focusing on training instances close to the critical region leads to more robust models that generalize to the overall distribution better than models trained via \emph{ad-hoc} sampling from the entire space).

\subsection{Grounded Rule Language (GRL)}
\label{gl}

The \emph{Grounded Rule Language} is a straightforward translation of the clauses in a random Boolean formula into grounded propositional (if-then) rules, similar to some of the rules used by \citet{clark2020transformers}. For example, as detailed in Figures~\ref{fig:lang_description} and~\ref{fig:lang_examples}, a clause with three literals: 
\begin{align}
\pm \mathbf{v}_{1} \lor \pm \mathbf{v}_{2} \lor \pm \mathbf{v}_{3}
\label{eq:GRL}
\end{align}
can be translated as \emph{If (no) $\mathbf{v}_{1}$ and (no) $\mathbf{v}_{3}$ then (no) $\mathbf{v}_{3}$} (using the standard rules of logical equivalence) where each variable $\mathbf{v}_{j}$ is subsequently replaced with an English count noun (i.e., any noun that can be made plural and made into a singular form with the determiner \emph{a}/\emph{an}). 


We choose a fixed set of around 50 nouns about food for our main \textbf{GRL} set reported in Table~\ref{tab:data_comparison} (see examples in Figure~\ref{fig:lang_examples}).  Each instance in our main set is characterized by a varying number of SAT variables or nouns, ranging from 5 to 12, which we discuss and motivate below.

We note that while the propositions in these fragments (e.g., \emph{carrot, steak}) deviate slightly from propositions encountered in ordinary language, one interpretation of the resulting theories is that they are akin to cooking recipes: e.g., \emph{if (you have) carrot and not steak then (you need to have) apples}. Figuring out whether the set of sentences is satisfiable is equivalent to deciding whether there is a coherent recipe underlying the rules. The decision to create data in a truncated form (i.e., without verbs) is due to the following considerations: some of the transformer models we probe have strict token limits which are easily exceeded when expressing the target hard computational problems using longer phrases; and leaving out this information does not affect the complexity of the resulting reasoning problem that we are interested in probing.\footnote{Similar arguments are used to justify compositional reasoning probing benchmarks, such as SCAN \cite{lake2018generalization}, which deviate even more sharply from ordinary natural language.} In the next section, we describe our second fragment that aims to capture more conventional linguistic constructions.




\subsection{Grounded Relative Clause Fragment (RCL)}
\label{sec:rc}

The grounded relative clause fragment is characterized by the relative clause rule construction \emph{Every $\mathbf{X}$ who is (not) a $\mathbf{Y}$ is (not) a $\mathbf{Z}$}, which, via its  translation from first-order logic: 
\begin{align}
\forall x.\ \ \mathbf{X}(x) \land \pm \mathbf{Y}(x) \to \pm \mathbf{Z}(y),
\label{eq:universal}
\end{align}
corresponds to boolean clauses of the form  $\neg \mathbf{v}_{1} \lor \pm \mathbf{v}_{2} \lor \pm \mathbf{v}_{3}$ containing up to two positive literals (where each variable corresponds to a count noun, or predicates $\mathbf{X},\mathbf{Y},\mathbf{Z}$). To allow for clauses with up to three positive literals, we add the rule template \emph{Everything that is (not) an $\mathbf{X}$ and (not) a $\mathbf{Y}$ is (not) a $\mathbf{Z}$}, where \emph{everything} universally quantifiers over the entire domain.

We obtain a mapping to propositional logic by assuming finite domains following some theoretical studies on quantifiers  \cite{westerstaahl1984some,szymanik2016quantifiers} and work on utilizing propositional logic for various reasoning problems in classical AI \cite{kautz1992planning,kautz1996encoding}. More specifically, grounding formulas such as Equation~\ref{eq:universal} relies on having clause translations \emph{$\mathbf{c}$ is (not) a/an $X$ or (not) a/an $Y$ or (not) a/an $Z$} that allow for introducing disjunctive facts that involve constants or proper nouns (denoted as $\mathbf{c}$); given a set of universally quantified rules and such disjunctive facts, all universals rules are expanded to group propositions over all constants out to arrive at a final grounded formula. 

Count and proper nouns are selected from a small inventory of noun types (as above, around 50) about people and their occupations (see again Figure~\ref{fig:lang_description}). A particular feature of this fragment is that through such universally quantified rules and their expansion to propositional logic, we can arrive at more complex reasoning problems that significantly increasing the number of \textbf{ground variables} and size of the target problems without dramatically increasing the size of the natural language input. The rules in our data are sampled from random 3SAT formulae over a fixed set of (5-8) variables and are coupled with an additional set of random clauses for disjunctive rule instances. While these resulting boolean formulas deviate from strict random 3SAT, the expansion of universal rules over a set of constants preserves the ratio of variables and clauses, which give rise to the same phase-change phenomena illustrated in Figure~\ref{fig:phase_trans}, allowing us to find the hard cases in the critical region. 




\subsection{Sampling Strategies and Proposed Datasets}
\label{sec:sampling-and-datasets}

A summary of our datasets is shown in Table~\ref{tab:data_comparison}, along with a comparison to the standard RuleTaker dataset converted to SAT.\footnote{More details about this conversion and technical details about the RuleTaker language can be found in the appendix.} As described above, the NLSat instances  that constitute the grounded rule language (\textbf{GCL}$_{5,12}$) and the grounded relative clause fragment (\textbf{RCL}$_{16,70}$) are characterized by the number of variables contained in their underlying Boolean formulae (with \textbf{d}$_{\#vars}$ denoting the overall range), which are uniformly represented in each dataset to allow for later inspection of model performance. For each variable amount, the majority of Boolean formulae are sampled from the critical 0.5 ($\pm 0.1$) probability region by heuristically controlling the $(\alpha_{\min},\alpha_{\max})$ clause variable ratio range in Algorithm~\ref{alg:sampling} (henceforth, \textbf{hard} sampling\footnote{We also added a small number of problems around the critical region in training to encourage diversity.}), which we later show leads to advantages over to both \textbf{naive sampling} (i.e., choosing instances randomly within a large range)  and \textbf{biased sampling} (i.e., sampling \emph{easy} instances from the extreme ends of the phase change) strategies (see Figure~\ref{fig:sampling_results}). Formulae and their translations are then randomly split into train and evaluation sets using a  80\% (train) / 20\% (dev,test) ratio.

\begin{table}[ht]
    \centering 

    
    {\footnotesize
    \begin{tabular}{| p{1.2cm} p{.5cm} p{1.4cm}  p{1.2cm} p{1.5cm} |}
    \multicolumn{1}{c}{} & &  \multicolumn{3}{c}{\emph{Language complexity and SAT metrics}} \\ \hline  
    \textbf{Dataset} ($\textbf{d}_{\#vars}$) & \textbf{Size} & \textbf{Complexity} {\tiny (NP-complete?)} & \textbf{Conflicts {\tiny (avg\textcolor{red}{/}med.)}} &  \textbf{Decisions {\tiny (avg\textcolor{red}{/}med.)}}\\ \hline 
    \textbf{RuleTaker} & 130k & \multicolumn{1}{c}{yes} & 0.0,\textcolor{red}{/}0.0 & 6.6\textcolor{red}{/}0.0  \\ \cdashline{1-5}
    \textbf{GRL}$_{5,12}$ & 187k & \multicolumn{1}{c}{yes} & 3.4\textcolor{red}{/}4.0 & 5.4\textcolor{red}{/}4.0 \\ 
    \textbf{RCL}$_{16,70}$ & 219k & \multicolumn{1}{c}{yes} & 7.6\textcolor{red}{/}6.0 & 29.7\textcolor{red}{/}6.0 \\ \cdashline{1-5}
    \textbf{GRL}-eval$_{20,50}$ & 17k & \multicolumn{1}{c}{yes} & 22.0\textcolor{red}{/}13.0 & 29.3\textcolor{red}{/}13.0 \\ \hline 
    \end{tabular}}
    
    \vspace{.2cm}

    \caption{The RuleTaker dataset, while similar in terms of dataset \textbf{size} and formal language \textbf{complexity} as our rule language datasets (\textbf{GRL} and \textbf{RCL}), is substantially simpler in terms of two standard SAT-based empirical complexity metrics: number of \textbf{conflicts} and \textbf{decisions}.}
    \label{tab:data_comparison}
\end{table}

A particular advantage of having boolean formulae associated with our target data is that we can use automatic reasoning tools to obtain empirical measurements of problem hardness. Using the off-the-shell theorem prover Z3~\cite{de2008z3}, we report the average/median (\textbf{avg}\textcolor{red}{/}\textbf{med.}) number of \textbf{decisions} (e.g., number of variable assignments after pre-processing) and \textbf{conflicts} (amount of backtracking performed for obtaining more complex proofs) needed by its \emph{solve} method on each datasets. While such statistics are often tied to the internals of a solver (especially \#\emph{decisions}), there still are some notable observations.

We see from Table~\ref{tab:data_comparison} that RuleTaker, in spite of its language's high theoretical complexity, is limited to the simplest forms of deductive inference as evidenced by having very few problems involving any conflicts and decisions at all (the median number for both is $0$). In sharp contrast, our new datasets, via our \textbf{hard} sampling strategy, offer a much wider range of problem difficulty. By retrofitting our randomly sampled 3SAT formula to include theories with 2SAT and single propositions similar to RuleTaker theories, we are also able to construct a substantially harder RuleTaker dataset (model performance to be discussed in Table~\ref{tab:hard-RT}).\footnote{Such retrofitting involves modifying Algorithm~\ref{alg:sampling} to allow for sampling of clause variables \emph{with replacement} (in \textbf{line 7}) which yields clauses with repeated variables that we convert into 2SAT and units. More details are included in the appendix.}

An important caveat is that while our new problems are of a higher degree of difficulty compared with problems in datasets like RuleTaker, they are still of vastly low complexity (both in terms of number of variables and the statistics shown in Table~\ref{tab:data_comparison}) compared to the much harder SAT instances encountered in the mainstream SAT literature \cite{jarvisalo2012international}. The decision to limit problems to the number of variables we did (e.g., to a maximum of 12 variables for \textbf{GRL}$_{5,12}$), is partly practical, and due to considerations such as token limits in the models we describe next and overall training efficiency. As we will see, these problems, while simple for mainstream SAT solvers, are still quite challenging for state-of-the-art transformer models and thus valuable for advancing research on the latter. Following \citet{reed2015neural}, the decision also reflects the idea that we should aim to train models that can perform \emph{scale-invariant} reasoning by generalizing from small problems. For this purpose, we create an additional held-out set of considerably larger grounded rule reasoning problems \textbf{GRL}-eval$_{20,50}$ to measure scale-invariance.





\section{Experimental Setup}

\paragraph{Task Definition.} Formally, a NLSat dataset $D = \{ (p^{(d)}, l^{(d))}  \}_{d}^{| D |}$ consists of NLSat problems $p$ (i.e., a set of rules expressed in natural language) paired with a label $l \in \{\mathbf{sat},\mathbf{unsat}\}$. The goal is to correctly predict the label (indicating satisfiability or not), thereby reducing to binary classification as in \citet{clark2020transformers}. 



\paragraph{Models.} Following recent studies on rule reasoning \cite{tafjord2020proofwriter}, our investigation centers around the pre-trained text-to-text transformer \textbf{T5-large} model (with around $770M$ parameters)\footnote{At an earlier iteration we also performed experiments T5-11b and found comparable performance. We note that a particular appeal of T5 is its use of \emph{relative} positional embeddings which allows us to evaluate on larger problems such as those in our \textbf{GRL}-eval set that exceed the 512 token limit from pre-training.} \cite{raffel2019exploring}. We also compare against \textbf{RoBERTa} (with around $355M$ parameters) \cite{liu2019roberta}. In each case we use the implementation from \citet{wolf2019huggingface}. Standardly, models are \emph{fine-tuned} to generate the target labels by optimizing for the cross-entropy loss over the target \textbf{sat} and \textbf{unsat} tokens or labels. Also standardly, model selection is by performed by doing a random search (in the style of \citet{devlin2018bert}) over target hyper-parameters (focusing especially on learning rate, random seed, and \# training iterations) and selecting models with the highest dev.\ score. As mentioned above, we also found intermediate pre-training on 60k simpler 2SAT instances (i.e., instances sampled with $p_{\textrm{int}} = 0$ in Algorithm~\ref{alg:sampling} with simpler natural language rule templates containing only 2 propositions) to be indispensable for stabilizing and improving model training efficiency on our main tasks.

\begin{table*}[t]
    \centering
    
    
    \vspace{.3cm}
    \resizebox{17.4cm}{!}{%
    {\footnotesize 
    \begin{tabular}{| l | c c  c c | c c c c |}
         \multicolumn{9}{c}{Dev \emph{Accuracy} \% for \textbf{easy} / \textbf{hard} (\colorbox{Gray}{i.i.d} and \colorbox{red!20}{o.o.d}) instances}   \\ 
        \cline{1-9}
        \multicolumn{1}{|c}{} & \multicolumn{4}{c}{main \textbf{GRL} splits (5 to 12 variables problems)} & \multicolumn{4}{c|}{\textbf{GRL}-eval (20-50 variables)} \\ \hline 
        \textbf{Model}$_{\emph{num\_var}}$ & \textbf{5var} &  \textbf{8var} & \textbf{10var} & \textbf{12var} & \textbf{20var} & \textbf{30var} & \textbf{40var} & \textbf{50var} \\ \hline
        T5$_{\emph{5\_var}}$ & 97.5 / \colorbox{Gray}{95.9} & 89.0 / 83.8 & 83.3 / 75.4 & 61.5 / 67.4 & 65.6 / 60.3 & 59.7 / 53.5 & 50.8 / 50.2 & 50.0 / 50.0 \\
        T5$_{\emph{8\_var}}$ & 96.2 / 94.0 & 92.4 / \colorbox{Gray}{87.9} & 87.7 / 81.6 & 73.6 / 74.8 & 74.4 / 67.5 & 67.1 / 58.3 & 53.5 / 51.2 & 50.1 / 50.0 \\ 
        T5$_{\emph{10\_var}}$ & 93.9 / 89.7 & 92.7 / 86.3 & 89.7 / \colorbox{Gray}{82.5} & 79.0 / 76.7 & 78.6 / 70.0 & 71.2 / 60.1 & 54.7 / 51.4 & 50.1 / 50.0 \\
        T5$_{\emph{12\_var}}$ & 94.5 / 91.1 & 91.5 / 84.9 & 87.7 / 80.7 & 77.3 / \colorbox{Gray}{81.0} & 77.8 / 70.1 & 70.7 / 60.3 & 53.3 / 51.4 & 50.0 / 50.0 \\ \cdashline{1-9}
        T5$_{\emph{5,12}}$ & 98.6 / \colorbox{Gray}{98.1} & 96.0 / \colorbox{Gray}{93.6} & 92.6 / \colorbox{Gray}{89.6} & 85.0 / \colorbox{Gray}{88.5} & 86.5 / \colorbox{red!20}{80.7} & 84.9 / \colorbox{red!20}{72.7} & 69.8 / \colorbox{red!20}{61.4} & 59.1 / \colorbox{red!20}{51.8} \\ \hline 
    \end{tabular}}}
    \caption{Models exhibit limited generalization. Performance (dev) of models trained on \textbf{GRL} problems containing differing numbers of a certain size and evaluated on \emph{easy} and \emph{hard} cases also of varying size. LEFT: Generalization across different combinations of problem sizes for training and evaluation. RIGHT: Generalization to larger instances never seen during training.}
    \label{tab:generalization_results}
\end{table*}

\begin{table}
    \centering
    \setlength\tabcolsep{1ex}
    {\footnotesize
    \begin{tabular}{| l c c c c c c |}
        \multicolumn{1}{c}{} & \multicolumn{6}{c}{Grounded Rule Language \textbf{GRL}, Accuracy\%} \\ \hline 
         \textbf{Model} & \textbf{5var} & \textbf{7var} & \textbf{8var} & \textbf{10var} & \textbf{12var} & \textbf{Avg.}  \\ \hline 
         \textbf{Random} & 50.0 & 50.0 & 50.0 & 50.0 & 50.0 & 50.0 \\ \cdashline{1-7}  
         \textbf{T5}$_{5,12}$ & 98.0 & 95.4 & 94.3 & 90.7 & 88.3 & 93.4 \\ 
         \textbf{RoBERTa}$_{5,12}$ & 96.4 & 92.0 & 90.2 & 85.4 & 83.4 & 89.5 \\ \hline 
    \end{tabular}}
    
    \vspace{.2cm}
    {\footnotesize
    \begin{tabular}{| l c c c c c |}
         \multicolumn{6}{c}{Grounded Relative Clause Language \textbf{RCL}, Accuracy\%} \\ \hline 
         \textbf{Model} & \textbf{16,21v} & \textbf{25,32v} & \textbf{35,48v} & \textbf{60,70v} & \textbf{Avg.}  \\ \hline 
         \textbf{Random} & 50.0 & 50.0 & 50.0 & 51.2 & 50.3 \\ \cdashline{1-6}
         \textbf{T5}$_{16,70}$ & 95.9  & 95.3 & 94.7 & 92.9 & 94.7 \\ 
         \textbf{RoBERTa}$_{16,70}$ & 96.0  & 95.9 & 94.9 & 94.0 & 95.2 \\ \hline 
    \end{tabular}}
    
    \caption{Models trained on \textbf{hard} sets are surprisingly good at some hard tasks in the i.i.d.\ setting. Performance (test) of models on the \textbf{GRL} and \textbf{RCL} fragments, split into performance on problems with differing number of variables.}
    \label{tab:test_results}
\end{table}

\begin{figure}
    \centering
    \includegraphics[scale=.43]{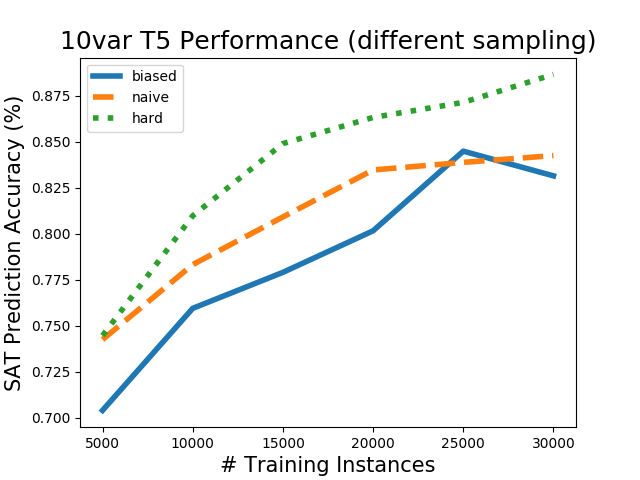}
    \vspace{.2cm}
    
    \centering 
    \begin{tabular}{| l | c c| }
        \multicolumn{1}{c}{} & \multicolumn{2}{c}{Accuracy\%} \\ \hline
         \textbf{Model} (\emph{sampling}) & \textbf{easy}$_{5,10}$ & \textbf{hard}$_{5,10}$  \\ \hline 
         \textbf{GRL}$_{10}$ (T5) (\emph{biased}) & 88.4 & 77.1 \\
         \textbf{GRL}$_{10}$ (T5) (\emph{naive}) & 89.7 & 78.7 \\ \cdashline{1-3}
         \textbf{GRL}$_{10}$ (T5) (\emph{hard}) & 92.4 & 86.4 \\ \hline 
    \end{tabular}
    \caption{Training on \textbf{hard} problems (our proposal) is much more effective than training on problems sampled in a \textbf{naive} or \textbf{biased} manner. TOP: Comparison of 10 variable model trained using different sampling strategies and tested across the \emph{full} distribution of \textbf{hard} (problems in critical region) and \textbf{easy} (problems at extreme of distribution) $5$ to $10$ variable problems (dev). BOTTOM: performance of the same 10 variable models on these different categories of problems.}
    \label{fig:sampling_results}
\end{figure}

\paragraph{Evaluation.}
We train models separately on our two languages (\textbf{GRL} and \textbf{RCL}) and their respective datasets (see again Table~\ref{tab:data_comparison}) in the manner described above. We report accuracy across sub-samples of evaluation data characterized by varying numbers of variables (i.e., the \textbf{Xvar} column in Tables~\ref{tab:generalization_results}-\ref{tab:test_results}). To better understand model generalization, we also experiment with training on small samples of data with a different number of variables for \textbf{GRL} as well as evaluation on a larger held-out \textbf{GRL}-eval set and \emph{easy} and \emph{hard} instances, as shown in Table~\ref{tab:generalization_results}. To better understand how different sampling strategies affect model performance, we perform experiments that measure the effect of different sampling strategies as shown in Figure~\ref{fig:sampling_results}.

Lastly, to verify the difficulty of our tasks, we also experimented a non-pretrained biLSTM encoder model implemented using AllenNLP~\cite{gardner2018allennlp}. While  not shown in the tables, we found, consistent with the results of \citet{clark2020transformers}, that such models perform near random chance.\footnote{As a check, we also verified that the same models obtained comparable results to the biLSTM baselines reported by \citet{clark2020transformers} on the original RuleTaker dataset.}




\section{Results and Findings}

Given our new set of hard algorithmic datasets, we aim to answer the following general question: \emph{How well can our main transformer models solve these types of \emph{hard} deductive reasoning problems?} As we describe in this section, while transformers perform well on some portion of our tasks and exhibit some degree of generalization, they still seem far from implementing the underlying algorithms needed for robust deductive reasoning. We also emphasize the following subtle point: knowing whether a model effectively solves a particular algorithmic problem or probing task such as SAT critically relies on understanding and specifying the target problem distribution that is being used for model development.\footnote{Such is also a lesson from the literature on hard SAT. Quoting \citet{mitchell1996some}: \emph{Random formulas have been used by many researchers to empirically evaluate the performance of SAT testing programs. The value of such studies depends upon careful selection of formula distribution... When using random formulas, an extensive enough study of the distribution's parameter space must be carried out ... if the results are to be meaningful}.} 


\paragraph{Effective sampling strategies are important for training robustness and reliable evaluation.} To assess the effectiveness of our \emph{hard} sampling strategy based on random 3SAT, we performed a smaller-scale experiment that involves training 10var problems in the \textbf{GRL} language, as summarized in Figure~\ref{fig:sampling_results}. Here we see that selecting training instances from the critical regions in 3SAT phase-changes (i.e., our default \textbf{hard} sampling strategy) allows models to generalize to the overall problem distribution and across other variable sizes (top figure). In contrast,  \textbf{naive}ly selecting from random parts of the distribution is not much better than selecting only from extreme (easy) ends of the distribution (\textbf{biased} sampling), and both lead to models not performing as well (8 points lower) on the hard subset of the evaluation set. Notably, the closeness of results between naive and biased sampling suggests that selecting deductive reasoning data in an \emph{ad hoc} fashion might often give rise to certain biases that harm robustness in a way similar to entirely biased sampling.

\begin{table}[ht]
    \centering 
    {\footnotesize
    \begin{tabular}{| l | ccc |}
         \multicolumn{1}{c}{} & \multicolumn{3}{c}{\emph{Model accuracy (\%)}} \\ \hline  
         \multicolumn{1}{|c|}{\emph{evaluation}}    &  Majority & \textbf{RT-}T5 & \textbf{RT-}RoBERTa \\ \hline 
         \textbf{RuleTaker} (RT) & \multirow{2}{*}{43.0} & \multirow{2}{*}{97.5} & \multirow{2}{*}{98.7} \\
         \ \ (standard i.i.d.) & & & \\ 
         \textbf{Hard RT} & \multirow{2}{*}{50.0} & \multirow{2}{*}{57.7} & \multirow{2}{*}{59.6} \\
         \ \ (our methodology) & & & \\ \hline
    \end{tabular}}
    \caption{While RT models excel on standard RuleTaker evaluation, they perform close to random on a Hard RT challenge set constructed using our methodology.}
\label{tab:hard-RT}
\end{table}

The importance of sampling for having reliable evaluations is further revealed in our experiments on sampling hard RuleTaker evaluation data from hard SAT, as shown in Table~\ref{tab:hard-RT}. While it is unclear what the exact distribution of problems defined in the RuleTaker domain is, the results in this table clearly demonstrate the efficacy of our general sampling framework in identifying hard datasets. They also show that even small changes in problem difficulty (namely, the relatively modest increase in standard empirical hardness measures, \#conflicts and \#decisions, as seen earlier in Table~\ref{tab:data_comparison}) can lead to dramatic differences in performance on existing benchmarks (e.g., a 39 point drop in performance for RoBERTa). This is reinforced by the differences in results between the \emph{easy/hard} problems shown in Table~\ref{tab:generalization_results} (e.g., 80\% vs. 71\% (avg) performance difference between easy/hard  20-40 variable problems for the full $T5_{5,12}$ model). Thus, without a proper understanding of the full distribution of target problems, it is often easy to draw inaccurate general conclusions about model capability by inadvertently focusing on easy instances.



\paragraph{Models trained on \emph{hard} sets can solve some hard tasks.} When looking at results on the hard instances (Table~\ref{tab:test_results}) we see that models trained on large collections of various types (i.e., on 150k-160k instances, see again Table~\ref{tab:data_comparison}) far outpace our baselines and achieve high performance on problems with not too many variables (e.g., \textbf{5}-\textbf{7} variables problems for \textbf{GRL} and \textbf{16}-\textbf{48} variables problems for \textbf{RCL}). A particularly intriguing result is the higher performance of models on the \textbf{RCL} language (with around 93-94\% accuracy on problems with 60-70 \emph{ground} variables) which was designed to be more complex by having quantified rules and constants that expand out to a much larger set of boolean variables. Given that the underlying rules were constructed from random 3SAT formula with a relatively smaller set of variables (5-8), this suggests that the model is able to learn some form of symmetry between the underlying rules and the instantiated rule propositions related to constants. 

\paragraph{Models exhibit limited generalization.}
Less impressive results are shown in the Table~\ref{tab:generalization_results}, where we see that models trained on small variable problems and fewer data fail to generalize to larger problems (e.g., generalizing from 5 variables to 10 or 12 variables). More strikingly, we see that even our best models fail to solve the \textbf{GPL}-eval evaluation task; while this is not altogether surprising, it suggests that state-of-the-art transformer models are still far from learning the underlying algorithms associated with deductive inference. 
 


\section{Closing Remarks}

With the advent of increasingly larger pre-trained models, including those that now allow for processing of tens of thousands of tokens \cite{beltagy2020longformer}, understanding the limits of how much aggregation of information over text models are capable of is an important area of study. Given that the type of algorithmic tasks we study in this paper are concerned with the most complex forms of information aggregation, we believe that our results can bear on these bigger issues about model design. When optimized for problem hardness, we see that models on our datasets still exhibit little ability to generalize in a scale-invariant fashion that is required for effectively generalizing their reasoning abilities to larger problems. Moving forward, we believe that new modeling approaches and architectures (e.g., ones that focus on problem decomposition \cite{andreas2016learning,khot2020text}) might be a fruitful avenue, which we believe our new algorithmic tasks and sampling strategies for finding hard datasets can assist in exploring. 

\section{Acknowledgments}
We thank the members of the Aristo team at AI2 for their feedback at various stages of this work, in particular Peter Clark and Oyvind Tafjord, as well as the Beaker team (\url{https://beaker.org/}) for their assistance and support with experiments. Special thanks also to Gregor Betz and Christian Voigt for helpful discussions.

\bibliography{NLSat}

\appendix

\begin{figure*}
    \centering
    {\footnotesize
    \begin{tabular}{| p{6cm} | p{10cm} |}
        \hline 
         \textbf{Step 1}: Find random 3SAT instances with modification to Algorithm~\ref{alg:sampling} that allows for clauses with \textbf{repeat}ed variables (i.e., removing the \emph{uniqueness} constraint on line \textbf{7} to sample \emph{with replacement}) & 
            $(\mathbf{v}_{1} \lor \neg \mathbf{v}_{2} \lor \neg \mathbf{v}_{5}) \land (\neg \mathbf{v}_{2} \lor \neg \mathbf{v}_{3} \lor \neg \mathbf{v}_{4}) \land  \underbrace{(\neg \mathbf{v}_{5} \lor  \neg \mathbf{v}_{5} \lor \neg \mathbf{v}_{5})}_{\textbf{repeat (unit)}} \land \underbrace{(\mathbf{v}_{1} \lor  \mathbf{v}_{1} \lor \mathbf{v}_{1})}_{\textbf{repeat (unit)}} \land  (\mathbf{v}_{3} \lor \mathbf{v}_{4} \lor \neg \mathbf{v}_{2}) \land \newline (\neg \mathbf{v}_{2} \lor \neg \mathbf{v}_{3} \lor \neg \mathbf{v}_{1}) \land ... \land \underbrace{(\neg \mathbf{v}_{1} \lor \neg \mathbf{v}_{1} \lor \mathbf{v}_{3})}_{\textbf{repeat (2SAT)}}$ \\ \hline 
         \textbf{Step 2}  Remove repeats, split formula into \textbf{rules} (i.e., 2/3 SAT clauses) and \textbf{facts} (i.e., units); find problems whose \textbf{rules} are satisfiable. 
            & $\underbrace{(\mathbf{v}_{1} \lor \neg \mathbf{v}_{2} \lor \neg \mathbf{v}_{5}) \land (\neg \mathbf{v}_{2} \lor \neg \mathbf{v}_{3} \lor \neg \mathbf{v}_{4}) (\neg \mathbf{v}_{5} \lor  \neg \mathbf{v}_{5} \lor \neg \mathbf{v}_{5}) \land ... }_{\textbf{rules} (sat.)} \land$  $\underbrace{\neg \mathbf{v}_{5} \land \mathbf{v}_{1}}_{\textbf{facts}}$ \\ \hline 
         \textbf{Step 3} Translate \textbf{rules} and \textbf{facts} to English using the RuleTaker templates from Figure~\ref{fig:ruletaker_complexity}. Treat some facts as \textbf{conjectures}, or the queries to be proven given the \textbf{Rules} and \textbf{Facts}.  
            & \textbf{Rules: }\emph{If $\underbrace{\text{the lion is not red}}_{\neg \mathbf{v}_{1}}$ and $\underbrace{\text{the lion is round}}_{\mathbf{v}_{2}}$ then $\underbrace{\text{the lion is not green}}_{\neg \mathbf{v}_{5}}$. If the $\underbrace{\text{lion is round}}_{\mathbf{v}_{2}}$ and $\underbrace{\text{the lion is young}}_{\mathbf{v}_{3}}$ then $\underbrace{\text{the lion is not rough}}_{\neg\mathbf{v}_{4}}$...} \newline \textbf{Facts: } $\underbrace{\text{The lion is not green}}_{\neg \mathbf{v}_{5}}$... \textbf{Conjecture: } $\underbrace{\text{The lion is red}}_{\mathbf{v}_{1}}$. \\ \hline
    \end{tabular}}

    \centering
    \caption{An illustration of the \emph{retrofitting} algorithm used to find hard RuleTaker \textbf{theories} (rule and facts) from random 3SAT using a contrived example with grounded rules over 5 variables. }
    \label{fig:retrofitting}
\end{figure*} 

\begin{figure}
    \centering
    {\footnotesize
    \begin{tabular}{| l| p{5cm} |}
        \hline 
         \textbf{Ground Rules} &  If the $\mathbf{c}$ is (not) \hlgray{$\mathbf{X}$} then the $\mathbf{c}$ is (not) \hlgray{$\mathbf{Y}$}. \textcolor{red}{If the $\mathbf{c}$ is (not) \hlgray{$\mathbf{X}$} and the $\mathbf{c}$ is (not) \hlgray{$\mathbf{Y}$} then the $\mathbf{c}$ is (not) \hlgray{$\mathbf{Z}$}.}  \\ \hline 
         \textbf{Quantified Rules} & If something is \hlgray{$\mathbf{X}$} and (not) \hlgray{$\mathbf{Y}$} then it is (not) \hlgray{$\mathbf{Z}$}. If something is (not) \hlgray{$\mathbf{X}$} then it is (not) \hlgray{$\mathbf{Y}$}. All \hlgray{$\mathbf{X}$}, \hlgray{$\mathbf{Y}$} things are (not) \hlgray{$\mathbf{Z}$}  \\  \hline 
    \end{tabular}}
    \caption{A subset of the rule templates encountered in the original RuleTaker language from \citet{clark2020transformers}.} 
    \label{fig:ruletaker_complexity}
\end{figure}

\section{Appendix}

\subsection{RuleTaker Details}
\label{subsec:ruletaker_details}


\paragraph{Complexity of RuleTaker Language} The rules in the original RuleTaker language \cite{clark2020transformers} take two general forms: \textbf{grounded rules} and \textbf{quantified rules}, a subset of which is shown in Figure~\ref{fig:ruletaker_complexity}. To demonstrate the NP-completeness of the RuleTaker language, it suffices to show that an arbitrary 3SAT formula $F$ can be expressed in this rule language such that $F$ is satisfiable if and only if the resulting RuleTaker theory is satisfiable (under propositional semantics). To this end, we observe that there are $4$ distinct atomic forms of a 3SAT clause corresponding to 0, 1, 2, or 3 positive literals (after accounting for logical equivalences obtained via the commutativity of disjunction). All of these atomic forms, denoted $\pm\mathbf{X} \lor \pm \mathbf{Y} \lor \pm \mathbf{Z}$ (where $\pm$ indicates the literal may be positive \emph{or} negated), can be represented by one of the aforementioned grounded rules, with appropriately placed \emph{not} modifiers:
\begin{figure}[ht]
\centering
{\footnotesize
\begin{tabular}{| l l |}
    \hline 
    \multicolumn{1}{|c}{\textbf{clausal form}} & \multicolumn{1}{c|}{\textbf{rule translation}} \\ \hline 
    $\pm\mathbf{X} \lor \pm \mathbf{Y} \lor \pm \mathbf{Z}$ & \emph{If the c is (not) $\mathbf{X}$ and the c is (not)} \\ 
    & \emph{$\mathbf{Y}$ then the c is (not) $\mathbf{Z}$.} \\ \hline
\end{tabular}}
\end{figure}

The original RuleTaker dataset includes instantiations of the above single rule that cover the $4$ distinct atomic forms.\footnote{Some corresponding rules from the original dataset: \emph{If the tiger is not big and the tiger is not blue then the tiger is cold, If the mouse is kind and the mouse is green then the mouse is blue, If the tiger is young and the tiger is big then the tiger is not blue, If the tiger is not blue and the tiger is not young then the tiger is not green.}} We note that a similar argument can be made for proving the NP-completeness of the other languages used in this work. 

\paragraph{Retrofitting random 3SAT to RuleTaker Theories} An example of how we \emph{retrofit} random 3SAT to create hard RuleTaker instances is shown in Figure~\ref{fig:retrofitting}. Given that RuleTaker theories (see again the example in Figure~\ref{fig:top_fig}) includes both \emph{2SAT clauses} (i.e., rules corresponding to clauses with two propositions, e.g., \emph{If the lion is red then it is rough}, in clausal form: $\neg A \lor B$) and \emph{units} (i.e., clauses with single propositions, e.g., \emph{The lion is red} or $A$), a particular difficulty is converting random 3SAT formula (where each clause contains exactly 3 propositions, e.g., $A \lor B \lor C$) to such forms.

Our idea is to modify Algorithm~\ref{alg:sampling} to allow for repeated clause variables that we can subsequently convert to 2SAT and units; technically this amounts to altering \textbf{line 7} to allow for \emph{sampling with replacement} such that we can produce clauses of the following form: $A \lor A \lor A$ that we can convert to facts (e.g., \emph{The lion is red}). As in the ordinary application of Algorithm~\ref{alg:sampling}, such a procedure can be performed to produce boolean formulae containing a differing number of variables. To keep the problems of comparable size to the original RuleTaker, we created problems using a mixture of 5,6,7 boolean variables. As a consequence, these problems are still of relatively low complexity comparing to the types of reasoning problems we pursue in our new datasets. 


\paragraph{RuleTaker version} For our experiments, we use the open world assumption (OWA) version of RuleTaker from \citet{tafjord2020proofwriter}\footnote{Publicly available at: \url{https://allenai.org/data/proofwriter}. We trained models on the \emph{depth-3ext}, which was empirically shown to have high generalization across the different \emph{depth} reasoning tasks in \citet{clark2020transformers}.}. In contrast to the initial version of the dataset from \citet{clark2020transformers}, which makes a closed-world assumption (CWA) and is limited to two-way entailment classification, the OWA include three classes: \text{\textbf{Yes} (entailment), \textbf{No} (contradiction), \textbf{Unknown}}. 

To verify the correctness of the semantics, we compared against a manual SAT-based and SMT-based implementation of the RuleTaker language, which is available at \url{https://github.com/allenai/language_fragments}. We found around 1\% mismatched labels between the official dataset due to apparent errors in the translation from the CWA dataset and performed experiments on the corrected version of the dataset.

\end{document}